\documentclass[10pt,twocolumn]{article}

\usepackage{cvpr}
\usepackage{times}
\usepackage{epsfig}
\usepackage{graphicx}
\usepackage{amsmath}
\usepackage{amssymb}
\usepackage{nameref}
\usepackage{subcaption}
\usepackage{float}
\usepackage{url}
\usepackage{color}
\usepackage{enumitem}
\usepackage{microtype}
\usepackage[subtle]{savetrees}

\usepackage[pagebackref=true,breaklinks=true,colorlinks,bookmarks=false]{hyperref}

\cvprfinalcopy 

\def\cvprPaperID{903} 

\iffalse
\newcommand {\kevin}[1]{{\color{cyan}\textbf{Kevin: }#1}\normalfont}
\newcommand {\clement}[1]{{\color{green}\textbf{Clem: }#1}\normalfont}
\newcommand {\matt}[1]{{\color{orange}\textbf{Matt: }#1}\normalfont}
\newcommand{\todo}[1]{\textcolor{red}{TODO: #1}\PackageWarning{TODO:}{#1!}}
\else
\newcommand {\kevin}[1]{}
\newcommand {\clement}[1]{}
\newcommand {\matt}[1]{}
\newcommand{\todo}[1]{}
\fi

\begin{document}

\title{Deep Burst Denoising}

\author{Cl\'ement~Godard\textsuperscript{1,*}\hspace{40pt}Kevin~Matzen\textsuperscript{2}\hspace{40pt}Matt~Uyttendaele\textsuperscript{2}\\
        \textsuperscript{1}University College London\hspace{40pt}\textsuperscript{2}Facebook}

\makeatletter
\g@addto@macro\@maketitle{
  \begin{figure}[H]
  \setlength{\linewidth}{\textwidth}
  \setlength{\hsize}{\textwidth}
  \centering        
  \includegraphics[width=\textwidth]{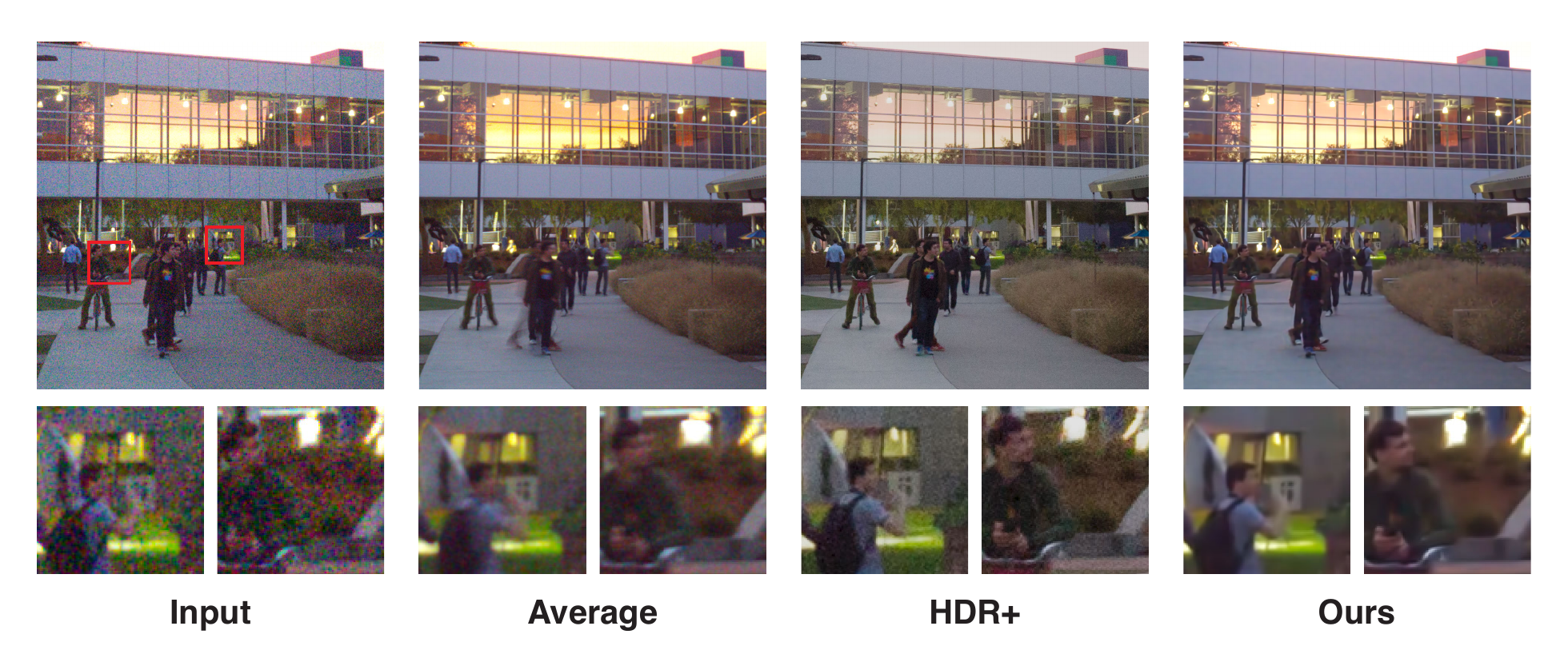}
    \caption{\textbf{Denoising on a real raw burst from \cite{hasinoff2016burst}.} Our method performs high levels of denoising on low-light bursts.}
    \label{fig:teaser}
\end{figure}
}
\makeatother

\maketitle
\renewcommand*{\thefootnote}{\fnsymbol{footnote}}
\setcounter{footnote}{1}
\footnotetext{This work was done during an internship at Facebook.}
\renewcommand*{\thefootnote}{\arabic{footnote}}
\setcounter{footnote}{0}

\begin{abstract}
Noise is an inherent issue of low-light image capture, one which is exacerbated on mobile devices due to their narrow apertures and small sensors. One strategy for mitigating noise in a low-light situation is to increase the shutter time of the camera, thus allowing each photosite to integrate more light and decrease noise variance. However, there are two downsides of long exposures: (a) bright regions can exceed the sensor range, and (b) camera and scene motion will result in blurred images. Another way of gathering more light is to capture multiple short (thus noisy) frames in a “burst” and intelligently integrate the content, thus avoiding the above downsides. In this paper, we use the burst-capture strategy and implement the intelligent integration via a recurrent fully convolutional deep neural net (CNN). We build our novel, multiframe architecture to be a simple addition to any single frame denoising model, and design to handle an arbitrary number of noisy input frames.
We show that it achieves state of the art denoising results on our burst dataset, improving on the best published multi-frame techniques, such as VBM4D \cite{maggioni2012video} and FlexISP \cite{heide2014flexisp}.
Finally, we explore other applications of image enhancement by integrating content from multiple frames and demonstrate that our DNN architecture generalizes well to image super-resolution.

\end{abstract}

\section{Introduction}

Noise reduction is one of the most important problems to solve in the design of an imaging pipeline. The most straight-forward solution is to collect as much light as possible when taking a photograph. This can be addressed in camera hardware through the use of a large aperture lens, sensors with large photosites, and high quality A/D conversion. However, relative to larger standalone cameras, e.g. a DSLR, modern smartphone cameras have compromised on each of these hardware elements. This makes noise much more of a problem in smartphone capture.

Another way to collect more light is to use a longer shutter time, allowing each photosite on the sensor to integrate light over a longer period of time. This is  commonly done by placing the camera on a tripod. The tripod is necessary as any motion of the camera will cause the collected light to blur across multiple photosites. This technique is limited though. First, any moving objects in the scene and residual camera motion will cause blur in the resulting photo. Second, the shutter time can only be set for as long as the brightest objects in the scene do not saturate the electron collecting capacity of a photosite. This means that for high dynamic range scenes, the darkest regions of the image may still exhibit significant noise while the brightest ones might staturate.  

In our method we also collect light over a longer period of time, by capturing a burst of photos. Burst photography addresses many of the issues above (a) it is available on inexpensive hardware, (b) it can capture moving subjects, and (c) it is less likely to suffer from blown-out highlights. In using a burst we make the design choice of leveraging a computational process to integrate light instead of a hardware process, such as in \cite{liu2014fast} and \cite{hasinoff2016burst}. In other words, we turn to computational photography.

Our computational process runs in several steps. First, the burst is stabilized by finding a homography for each frame that geometrically registers it to a common reference. Second, we employ a fully convolutional deep neural network (CNN) to denoise each frame individually. Third, we extend the CNN with a parallel recurrent network that integrates the information of all frames in the burst. 

The paper presents our work as follows. In section \ref{sec:related} we review previous single-frame and multi-frame denoising techniques. We also look at super-resolution, which can leverage multi-frame information. In section \ref{sec:method} we describe our recurrent network in detail and discuss training. In order to compare against previous work, the network is trained on simulated Gaussian noise. We also show that our solution works well when trained on Poisson distributed noise which is typical of a real-world imaging pipeline~\cite{conf/cvpr/HasinoffDF10}. 
In section \ref{sec:results}, we show significant increase in reconstruction quality on burst sequences in comparison to state of the art single-frame denoising and performance on par or better than recent state of the art multi-frame denoising methods. In addition we demonstrate that burst capture coupled with our recurrent network architecture generalizes well to super-resolution.

In summary our main contributions are:
\begin{itemize}[noitemsep]
  \item We introduce a recurrent "feature accumulator" network as a simple yet effective extension to single-frame denoising models, 
  \item Demonstrate that bursts provide a large improvement over the best deep learning based single-frame denoising techniques,
  \item Show that our model reaches performance on par with or better than recent state of the art multi-frame denoising methods, and
  \item Demonstrate that our recurrent architecture generalizes well to the related task of super-resolution.
\end{itemize}

\section{Related work}\label{sec:related}

This work addresses a variety of inverse problems, all of which can be formulated as consisting of (1) a target ``restored'' image, (2) a temporally-ordered set or ``burst'' of images, each of which is a corrupted observation of the target image, and (3) a function mapping the burst of images to the restored target. Such tasks include denoising and super-resolution.
Our goal is to craft this function, either through domain knowledge or through a data-driven approach, to solve these multi-image restoration problems.  

\subsection*{Denoising}
Data-driven single-image denoising research dates back to work that leverages block-level statistics within a single image.  One of the earliest works of this nature is Non-Local Means~\cite{buades2005non}, a method for taking a weighted average of blocks within an image based on similarity to a reference block. Dabov,~\etal~\cite{dabov2009bm3d} extend this concept of block-level filtering with a novel 3D filtering formulation.  This algorithm, BM3D, is the de facto method by which all other single-image methods are compared to today.

Learning-based methods have proliferated in the last few years.  These methods often make use of neural networks that are purely feed-forward~\cite{NIPS2012_4686,burger2012image,Zhang2017BeyondAG,NIPS2008_3506,gharbi2016deep,NIPS2013_5030,Zhang_2017_CVPR}, recurrent~\cite{ChenSPIE2016}, or a hybrid of the two~\cite{chen2017trainable}.  Methods such as Field of Experts~\cite{roth2005fields} have been shown to be successful in modeling natural image statistics for tasks such as denoising and inpainting with contrastive divergence.  Moreover, related tasks such as demosaicing and denoising have shown to benefit from joint formulations when posed in a learning framework~\cite{gharbi2016deep}. Finally, the recent work of \cite{chaitanya2017interactive} applied a recurrent architecture in the context of denoising ray-traced sequenced.

Multi-image variants of denoising methods exist and often focus on the best ways to align and combine images. Tico~\cite{tico2008multi} returns to a block-based paradigm, but this time, blocks ``within'' and ``across'' images in a burst can be used to produce a denoised estimate.  VBM3D~\cite{dabov2007image} and VBM4D ~\cite{maggioni2011video, maggioni2012video} provide extensions on top of the existing BM3D framework.  Liu,~\etal~\cite{liu2014fast} showed how similar denoising performance in terms of PSNR could be obtained in one tenth the time of VBM3D and one one-hundredth the time of VBM4D using a novel ``homography flow'' alignment scheme along with a ``consistent pixel'' compositing operator.  Systems such as FlexISP~\cite{heide2014flexisp} and ProxImaL~\cite{heide2016proximal} offer end-to-end formulations of the entire image processing pipeline, including demosaicing, alignment, deblurring, etc., which can be solved jointly through efficient optimization.

We in turn also make use of a deep model and base our CNN architecture on current state of the art single-frame methods \cite{remez2017deep, Zhang2017BeyondAG, Ledig_2017_CVPR}.

\subsection*{Super-Resolution}
Super-resolution is the task of taking one or more images of a fixed resolution as input and producing a fused or hallucinated image of higher resolution as output.

Nasrollahi,~\etal~\cite{nasrollahi2014super} offers a comprehensive survey of single-image super-resolution methods and Yang,~\etal~\cite{yang2014single} offers a benchmark and evaluation of several methods. Glasner,~\etal~\cite{Glasner2009} show that single images can be super-resolved without any need of an external database or prior by exploiting block-level statistics ``within'' the single image.  Other methods make use of sparse image statistics~\cite{yang2010image}.  Borman,~\etal offers a survey of multi-image methods~\cite{borman1998super}.  Farsiu,~\etal~\cite{farsiu2004fast} offers a fast and robust method for solving the multi-image super-resolution problem. More recently convolutional networks have shown very good results in single image super-resolution with the works of Dong~\etal~\cite{dong2016image} and the state of the art Ledig~\etal~\cite{Ledig_2017_CVPR}.

Our single-frame architecture takes inspiration by recent deep super-resolution models such as ~\cite{Ledig_2017_CVPR}.

\subsection{Neural Architectures}
It is worthwhile taking note that while image restoration approaches have been more often learning-based methods in recent years, there's also great diversity in how those learning problems are modeled.  In particular, neural network-based approaches have experienced a gradual progression in architectural sophistication over time.

In the work of Dong,~\etal~\cite{7115171}, a single, feed-forward CNN is used to super-resolve an input image.  This is a natural design as it leveraged what was then new advancements in discriminatively-trained neural networks designed for classification and applied them to a regression task.  The next step in architecture evolution was to use Recurrent Neural Networks, or RNNs, in place of the convolutional layers of the previous design.  The use of one or more RNNs in a network design can both be used to increase the effective depth and thus receptive field in a single-image network~\cite{ChenSPIE2016} or to integrate observations across many frames in a multi-image network.  Our work makes use of this latter principle.

While the introduction of RNNs led to network architectures with more effective depth and thus a larger receptive field with more context, the success of skip connections in classification networks~\cite{7780459} and segmentation networks~\cite{7478072,Ronneberger2015} motivated their use in restoration networks.  The work of Remez,~\etal~\cite{remez2017deep} illustrates this principle by computing additive noise predictions from each level of the network, which then sum to form the final noise prediction.  

We also make use of this concept, but rather than use skip connections directly, we extract activations from each level of our network which are then fed into corresponding RNNs for integration across all frames of a burst sequence.

\section{Method}\label{sec:method}

In this section we first identify a number of interesting goals we would like a multi-frame architecture to meet and then describe our method and how it achieves such goals. 

\subsection{Goals}\label{goals}
Our goal is to derive a method which, given a sequence of noisy images produces a denoised sequence.
We identified desirable properties, that a multi-frame denoising technique should satisfy:
\begin{enumerate}[noitemsep]
    \item {\bf Generalize to any number of frames.}  A single model should produce competitive results for any number of frames that it is given.
    \item {\bf Work for single-frame denoising.}  A corollary to the first criterion is that our method should be competitive for the single-frame case.
    \item {\bf Be robust to motion.}  Most real-world burst capture scenarios will exhibit both camera and scene motion.
    \item {\bf Denoise the entire sequence.}  Rather than simply denoise a single reference frame, as is the goal in most prior work, we aim to denoise the entire sequence, putting our goal closer to video denoising.
    \item {\bf Be temporally coherent.}  Denoising the entire sequence requires that we do not introduce flickering in the result.
    \item {\bf Generalize to a variety of image restoration tasks.} As discussed in Section~\ref{sec:related}, tasks such as super-resolution can benefit from image denoising methods, albeit, trained on different data.
\end{enumerate}

In the remainder of this section we will first describe a single-frame denoising model that produces competitive results with current state of the art models.  Then we will discuss how we extend this model to accommodate an arbitrary number of frames for multi-frame denoising and how it meets each of our goals.

\subsection{Single frame denoising}
We treat image denoising as a structured prediction problem, where the network is tasked with regressing a pixel-aligned denoised image $\tilde{I_s} = f_s(N, \theta_s)$ from noisy image $N$ with model parameters $\theta_s$. Following \cite{zhao2017loss} we train the network by minimizing the L1 distance between the predicted output and the ground-truth target image, $I$.
\begin{equation}
E_{\mathrm{SFD}} = | I - f_s(N, \theta_s) |
\end{equation}

To be competitive in the single-frame denoising scenario, and to meet our 2\textsuperscript{nd} goal, we take inspiration from the state of the art to derive an initial network architecture. Several existing architectures~\cite{Zhang2017BeyondAG,remez2017deep,Ledig_2017_CVPR} consist of the same base design: a fully convolutional architecture consisting of $L$ layers with $C$ channels each.

We therefore follow suit by choosing this simple architecture as our single frame denoising (SFD) base, with $L=8$, $C=64$, $3\times3$ convolutions and ReLU \cite{maas2013rectifier} activation functions, except on the last layer, as can be seen in Figure \ref{fig:architecture}. 

\subsection{Multi-frame denoising}

Following goals 2 and 4, we want our model to be competitive in the single-frame case while being able to denoise the entire input sequence.
Hence, given the set of all noisy images forming the sequence, $\{N^t\}$, we task the network to regress a denoised version of each noisy frame, $\tilde{I^t_m} = f^t_m(\{N^t\}, \theta_m)$ with model parameters $\theta_m$. Our complete training objecting is thus:

\begin{equation}
\begin{split}
E & = \sum_t^F E_{\mathrm{SFD}}^t + E_{\mathrm{MFD}}^t\\
& = \sum_t^F | I^t - f_s(N^t, \theta_s) | + | I^t - f^t_m(\{N^t\}, \theta_m) |
\end{split}
\end{equation}

\begin{figure}[t]
  \centering
    \includegraphics[width=0.99\linewidth, page=3]{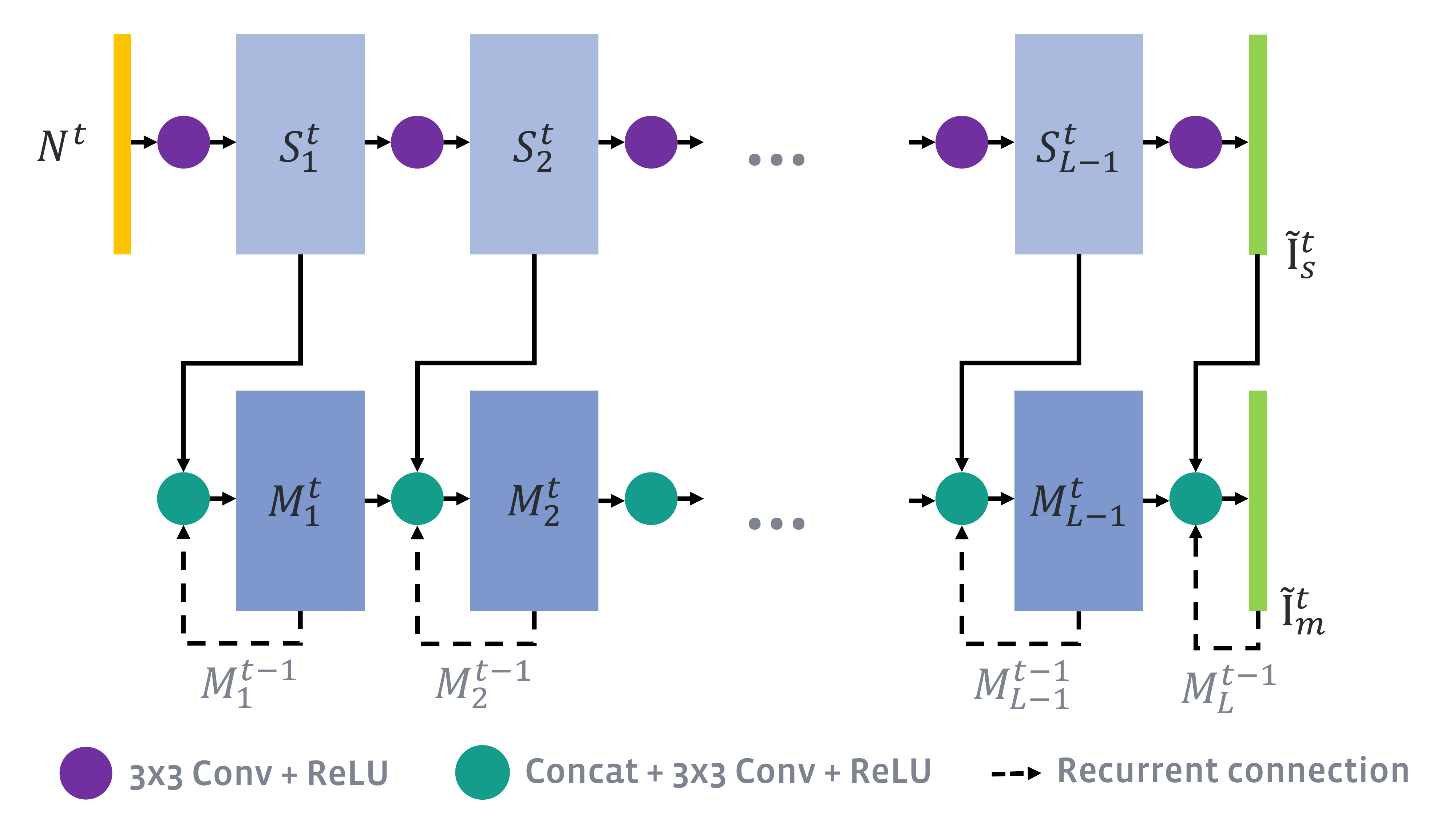}
    \caption{\textbf{Our recurrent denoising architecture.} The top part of our model is a single-frame denoiser (SFD, in light blue): it takes as input a noisy image $N^t$ and regresses a clean image $\tilde{I^t_s}$, its features $S_i^t$ are fed to the multi-frame denoiser (MFD, in darker blue) which also makes use of recurrent connections (in dotted lines) to output a clean image $\tilde{I^t_m}$.}
    \label{fig:architecture}
\end{figure}

A natural approach, which is already popular in the natural language and audio processing literature \cite{yin2017comparative}, is to process temporal data with recurrent neural networks (RNN) modules \cite{hopfield1982neural}. RNNs operate on sequences and maintain an internal state which is combined with the input at each time step. 
In our model, we make use of recurrent connections to aggregate activations produced by our SFD network for each frame, as we show in Figure~\ref{fig:architecture}. This allows for an arbitrary input sequence length, our first goal. Unlike \cite{chaitanya2017interactive} and \cite{wieschollek2017learning} which utilize a single-track network design, we use a two track network architecture with the top track dedicated to SFD and the bottom track dedicated to fusing those results into a final prediction for MFD.

By decoupling per-frame feature extraction from multi-frame aggregation, we enable the possibility for pre-training a network rapidly using only single-frame data.  In practice, we found that this pre-training not only accelerates the learning process, but also produces significantly better results in terms of PSNR than when we train the entire MFD from scratch.  The core intuition is that by first learning good features for SFD, we put the network in a good state for learning how to aggregate those features across observations, but still grant it the freedom to update those features by not freezing the SFD weights during training.

It is also important to note that the RNNs are connected in such a way as to permit the aggregation of observation features in several different ways.  Temporal connections within the RNNs help aggregate information ``across'' frames, but lateral connections ``within'' the MFD track permit the aggregation of information at different physical scales and at different levels of abstraction.

\section{Implementation and Results}\label{sec:results}

To show that our method fulfills the goals set in Section \ref{sec:method}, we evaluate it in multiple scenarios: single-image denoising, multi-frame denoising, and single-image super-resolution

\kevin{revisit later: We now evaluate our method in several steps. We first compare our single frame track denoiser with single frame denoising methods and show that much higher gains can be achieved through multi-frame methods.}

\subsection{Data}
We trained all the networks in our evaluation using a dataset consisting of Apple Live Photos. Live Photos are burst sequences captured by Apple iPhone 6S and above\footnote{\url{https://support.apple.com/en-us/HT207310}}. This dataset is very representative as it captures what mobile phone users like the photograph, and exhibits a wide range of scenes and motions.  Approximately 73k public sequences were scraped from a social media website with a resolution of $360\times480$.  We apply a burst stabilizer to each sequence, resulting in approximately 54.5k sequences successfully stabilized.  In Section~\ref{sec:stab} we describe our stabilization procedure in more detail.  50k sequences were used for training with an additional 3.5k reserved for validation and 1k reserved for testing.

\subsection{Stabilization}\label{sec:stab}
We implemented burst sequence stabilization using OpenCV\footnote{\url{https://opencv.org/}}.  In particular, we use a Lucas-Kanade tracker~\cite{Lucas:1981:IIR:1623264.1623280} to find correspondences between successive frames and then a rotation-only motion model and a static focal length guess to arrive at a homography for each frame.  We warp all frames of a sequence back into a reference frame's pose and crop and scale the sequence to maintain the original size and aspect ratio, but with the region of interest contained entirely within the valid regions of the warp. The stabilized sequences still exhibit some residual motion, either through moving objects or people, or through camera and scene motion which cannot be represented by a homography. This residual motion forces the network to adapt to non static scenes and be robust to motion, which is our 3\textsuperscript{rd} goal.

\subsection{Training details}
We implemented the neural network from Section~\ref{sec:method} using the Caffe2 framework\footnote{\url{https://caffe2.ai/}}.  Each model was trained using 4 Tesla M40 GPUs.  As described in Section~\ref{sec:method}, training took place in two stages.  First a single-frame model was trained.  This model used a batch size of 128 and was trained for 500 epochs in approximately 5 hours.  Using this single-frame model as initialization for the multi-frame (8-frame) model, we continue training with a batch size of 32 to accommodate the increased size of the multi-frame model over the single-frame model.  This second stage was trained for 125 epochs in approximately 20 hours.

We used Adam~\cite{Adam} with a learning rate of $10^{-4}$ which decays to zero following a square root law. We trained on $64\times64$ crops with random flips. Finally, we train the multi-frame model using back-propagation through time~\cite{werbos1988generalization}.

\subsection{Noise modelling}
We first evaluate our architecture using additive white Gaussian noise with $\sigma=15, 25, 50$ and $75$, in order to make comparison possible with previous methods, such as VBM4D. To be able to denoise real burst sequences, we modeled sensor noise following \cite{foi2009clipped} and trained separate models by adding Poisson noise, labelled \textit{a} in \cite{foi2009clipped}, with intensity ranging from 0.001 to 0.01 in linear space before converting back to sRGB and clipping. We also simulate Bayer filtering and reconstruct an RGB image using bilinear interpolation. Unless otherwise mentioned, we add synthetic noise \textit{before} stabilization.

\begin{figure}
  \centering
    \includegraphics[width=0.85\linewidth]{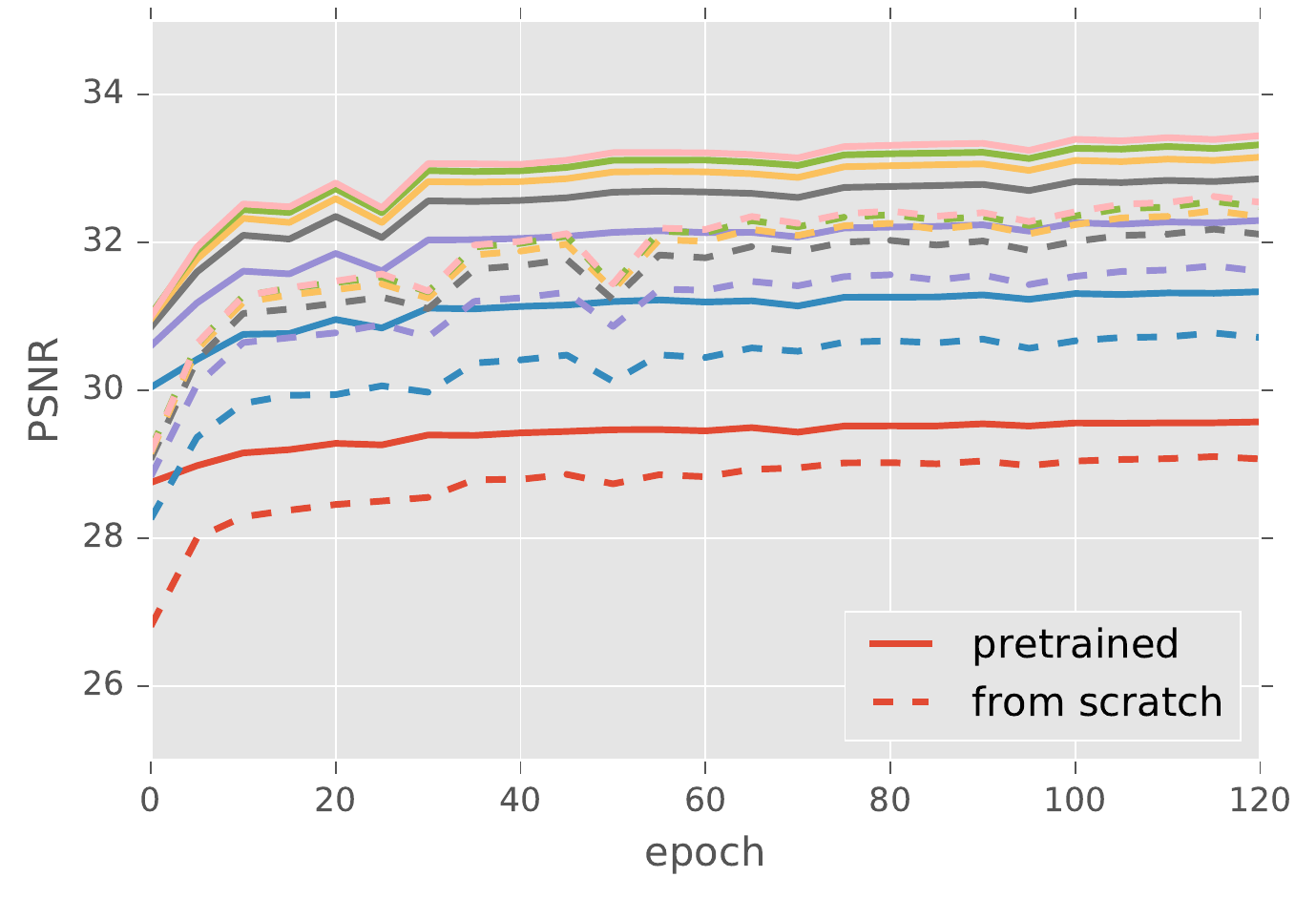}
    \caption{{\bf Effect of pre-training on multi-frame denoising with Gaussian noise $\sigma=50$.} Each color corresponds to the average PSNR of the frames in a sequence: $1^{st}$ (red), $2^{nd}$ (blue), $3^{rd}$ (purple), \etc. As we can see the pre-trained model shows a constant lead of 0.5dB over the model trained from scratch, and reaches a stable state much quicker.}
    \label{fig:pretraining}
\end{figure}

\subsection{Single frame denoising}

Here we compare our single frame denoiser with current state of the art methods on additive white Gaussian noise. We compare our own SFD, which is composed of 8 layers, with the two 20 layer networks of DenoiseNet (2017)~\cite{remez2017deep} and DnCNN (2017)~\cite{Zhang2017BeyondAG}. For the sake of comparison, we also include a 20 layer version of our SFD as well as reimplementations of both DnCNN and DenoiseNet. All models were trained for 2000 epochs on 8000 images from the PASCAL VOC2010~\cite{everingham2010pascal} using the training split from~\cite{remez2017deep}. We also include in the comparison BM3D (2009)~\cite{dabov2009bm3d} and TNRD (2015) ~\cite{chen2017trainable}.

All models were tested on BSD68~\cite{roth2005fields}, a set of 68 natural images from the Berkeley Segmentation Dataset~\cite{martin2001database}.
In Figure~\ref{tab:bsd68}, we can see diminishing returns in single frame denoising PSNR over the years, which confirms what Levin,~\etal~ describe in~\cite{levin2011natural}, despite the use of deep neural networks. We can see that our simpler SFD 20 layers model only slightly underperforms both DenoiseNet and DnCNN by $\sim0.2dB$. However, as we show in the following section, the PSNR gains brought by multi-frame processing vastly outshine fractional single frame PSNR improvements.

\begin{table}[]
\centering
\scalebox{0.8}{
\begin{tabular}{|l|c|c|c|c|}
\hline
                    & $\sigma=15$ & $\sigma=25$ & $\sigma=50$ & $\sigma=75$ \\ \hline
BM3D                & 31.10 & 28.57 & 25.62 & 24.20 \\ \hline
TNRD                & 31.41 & 28.91 & 25.95 &  -    \\ \hline
DenoiseNet \cite{remez2017deep} & 31.44 & 29.04 & 26.06 & 24.61 \\ \hline
DenoiseNet (reimpl) & 31.43 & 28.91 & 25.95 & 24.59 \\ \hline
DnCNN \cite{Zhang2017BeyondAG} & \textbf{31.73} & \textbf{29.23} & \textbf{26.23} &   -   \\ \hline
DnCNN (reimpl w/o BN)      & 31.42 & 28.86 & 25.99 & 24.30 \\ \hline \hline
SFD 8L              & 31.15 & 28.63 & 25.65 & 24.11 \\ \hline
SFD 20L             & 31.29 & 28.82 & 26.02 & 24.43 \\ \hline
\end{tabular}
}
\caption{\textbf{Single frame additive white Gaussian noise denoising comparison on BSD68 (PSNR).} Our simple SFD models match BM3D at 8 layers and get close to both DnCNN and DenoiseNet at 20 layers.}
\label{tab:bsd68}
\vspace{-3mm}
\end{table}

\subsection{Burst denoising}

\begin{figure*}
\vspace{-5mm}
  \centering
    \newcommand{\turnheightnew}{0.16\textwidth}

\resizebox{0.60\textwidth}{!}{\begin{tabular}{@{\hskip 1mm}c@{\hskip 1mm}c@{\hskip 1mm}c@{\hskip 1mm}c@{\hskip 1mm}c@{\hskip 1mm}c@{}}

Input & Average & VBM4D \cite{maggioni2011video} & SFD (Ours) & MFD (Ours) & Ground truth \\

\includegraphics[width=\turnheightnew, trim=6cm 4cm 13cm 8cm, clip=true]{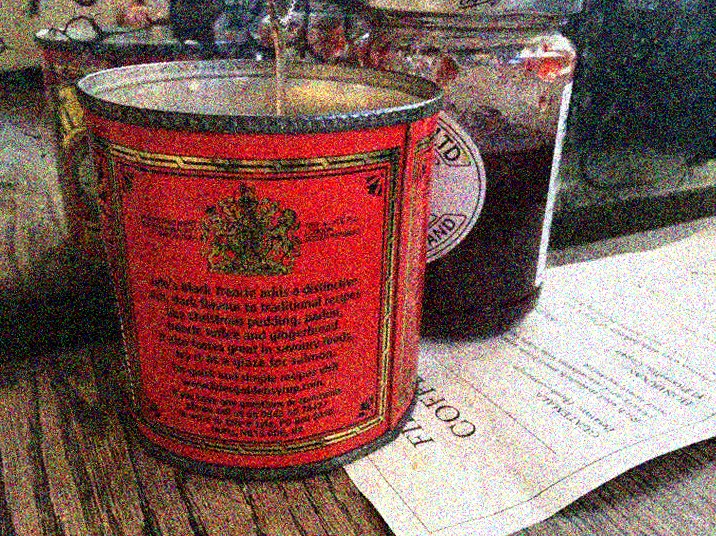} &
\includegraphics[width=\turnheightnew, trim=6cm 4cm 13cm 8cm, clip=true]{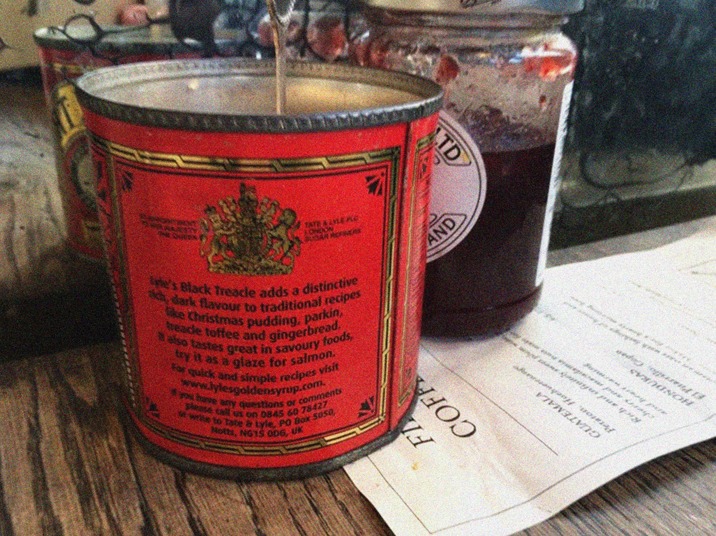} &
\includegraphics[width=\turnheightnew, trim=6cm 4cm 13cm 8cm, clip=true]{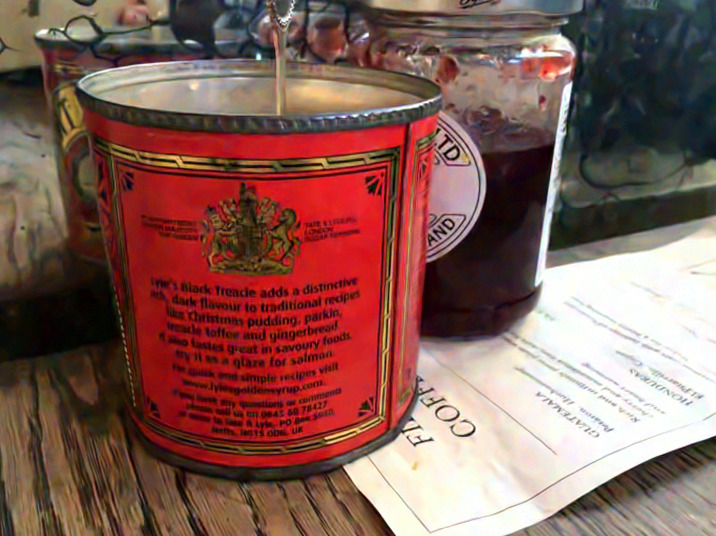} &
\includegraphics[width=\turnheightnew, trim=6cm 4cm 13cm 8cm, clip=true]{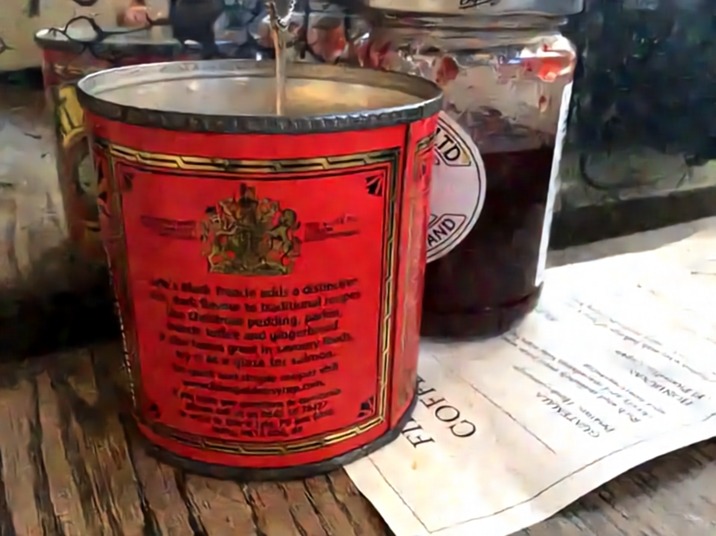} &
\includegraphics[width=\turnheightnew, trim=6cm 4cm 13cm 8cm, clip=true]{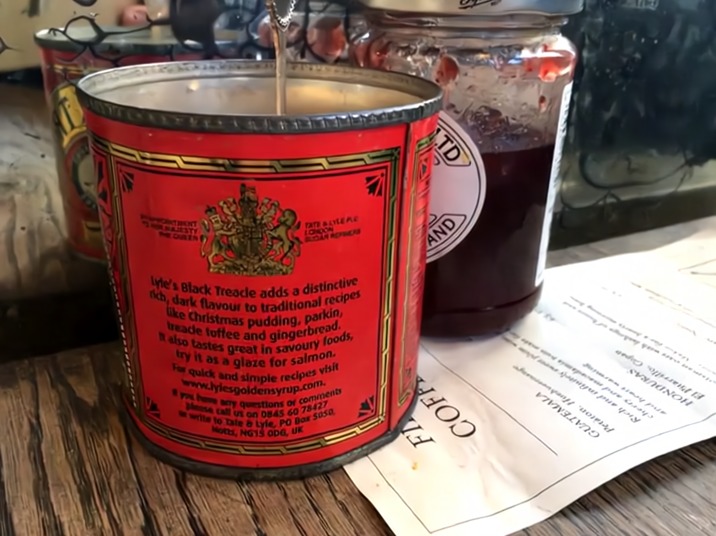} &
\includegraphics[width=\turnheightnew, trim=6cm 4cm 13cm 8cm, clip=true]{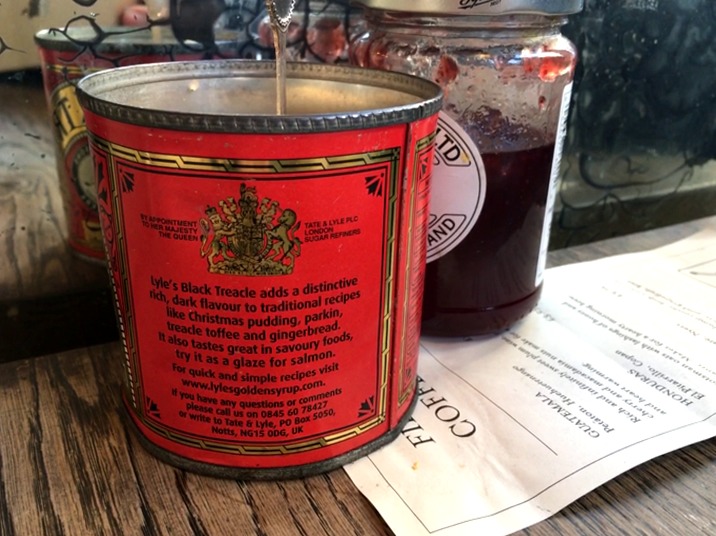}\\

\includegraphics[width=\turnheightnew, trim=6cm 4cm 13cm 7cm, clip=true]{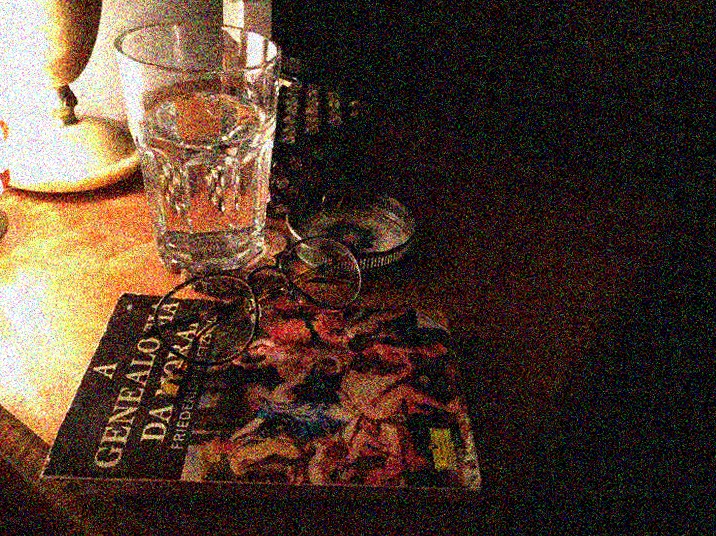} &
\includegraphics[width=\turnheightnew, trim=6cm 4cm 13cm 7cm, clip=true]{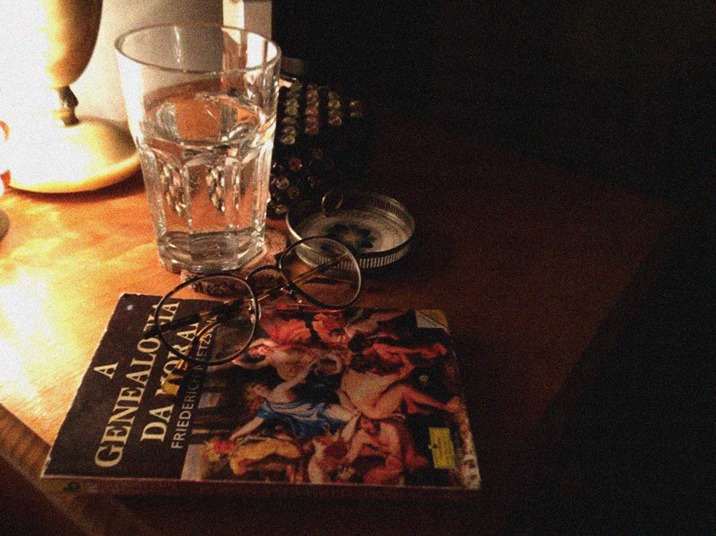} &
\includegraphics[width=\turnheightnew, trim=6cm 4cm 13cm 7cm, clip=true]{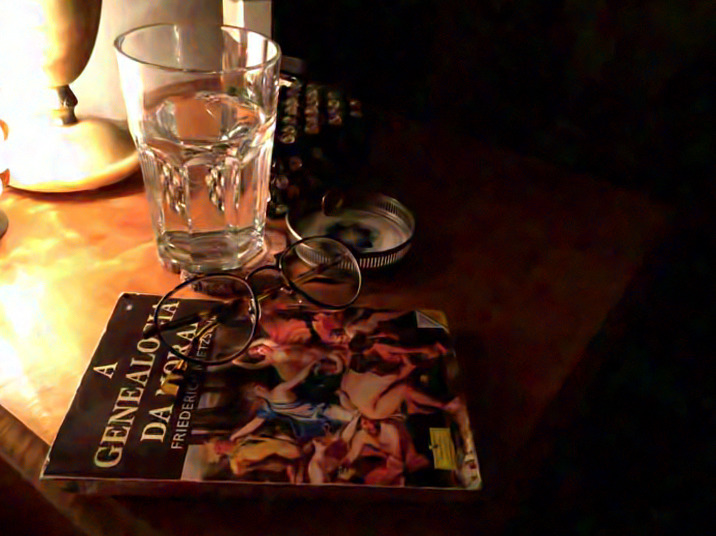} &
\includegraphics[width=\turnheightnew, trim=6cm 4cm 13cm 7cm, clip=true]{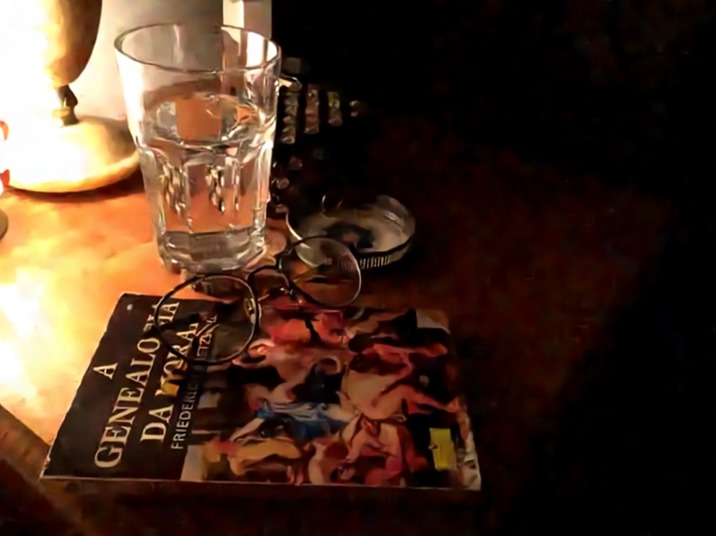} &
\includegraphics[width=\turnheightnew, trim=6cm 4cm 13cm 7cm, clip=true]{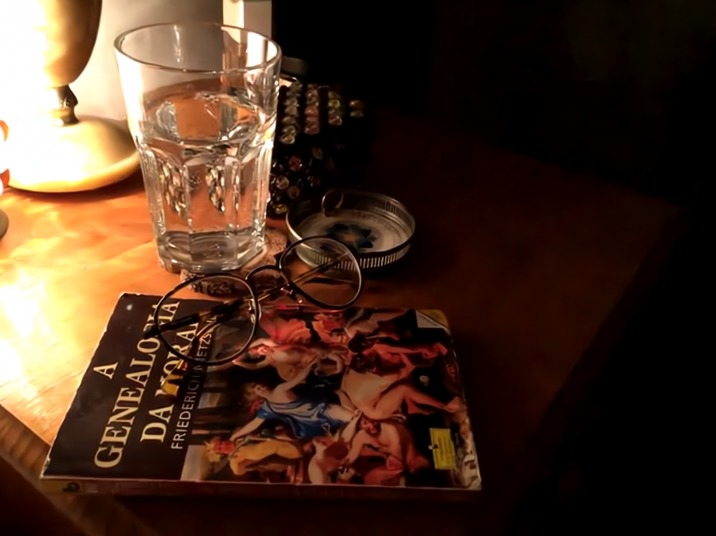} &
\includegraphics[width=\turnheightnew, trim=6cm 4cm 13cm 7cm, clip=true]{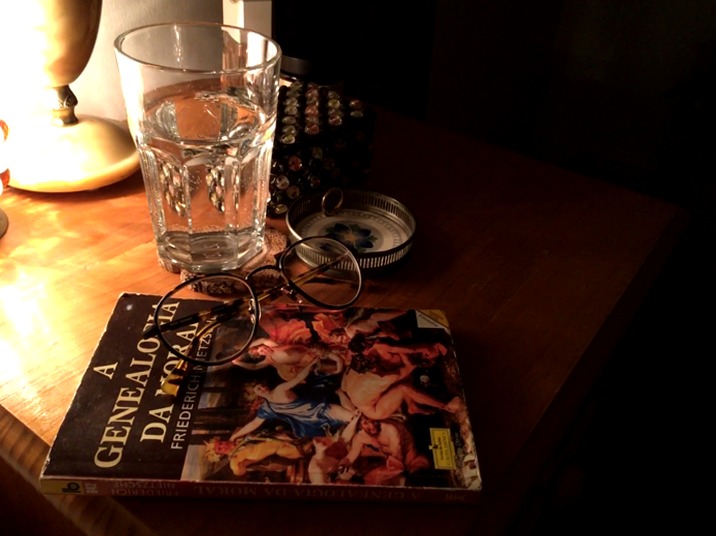}\\
\end{tabular}}
    \caption{\textbf{Multi-frame Gaussian denoising on stabilized Live Photo test data with $\sigma = 50$.} We can see that our MFD produces a significantly sharper image than both our SFD and VBM4D, the latter exhibiting significant temporal color flickering.}
    \label{fig:sigma_50}
\end{figure*}

\begin{figure*}
\centering
\includegraphics[width=0.65\textwidth]{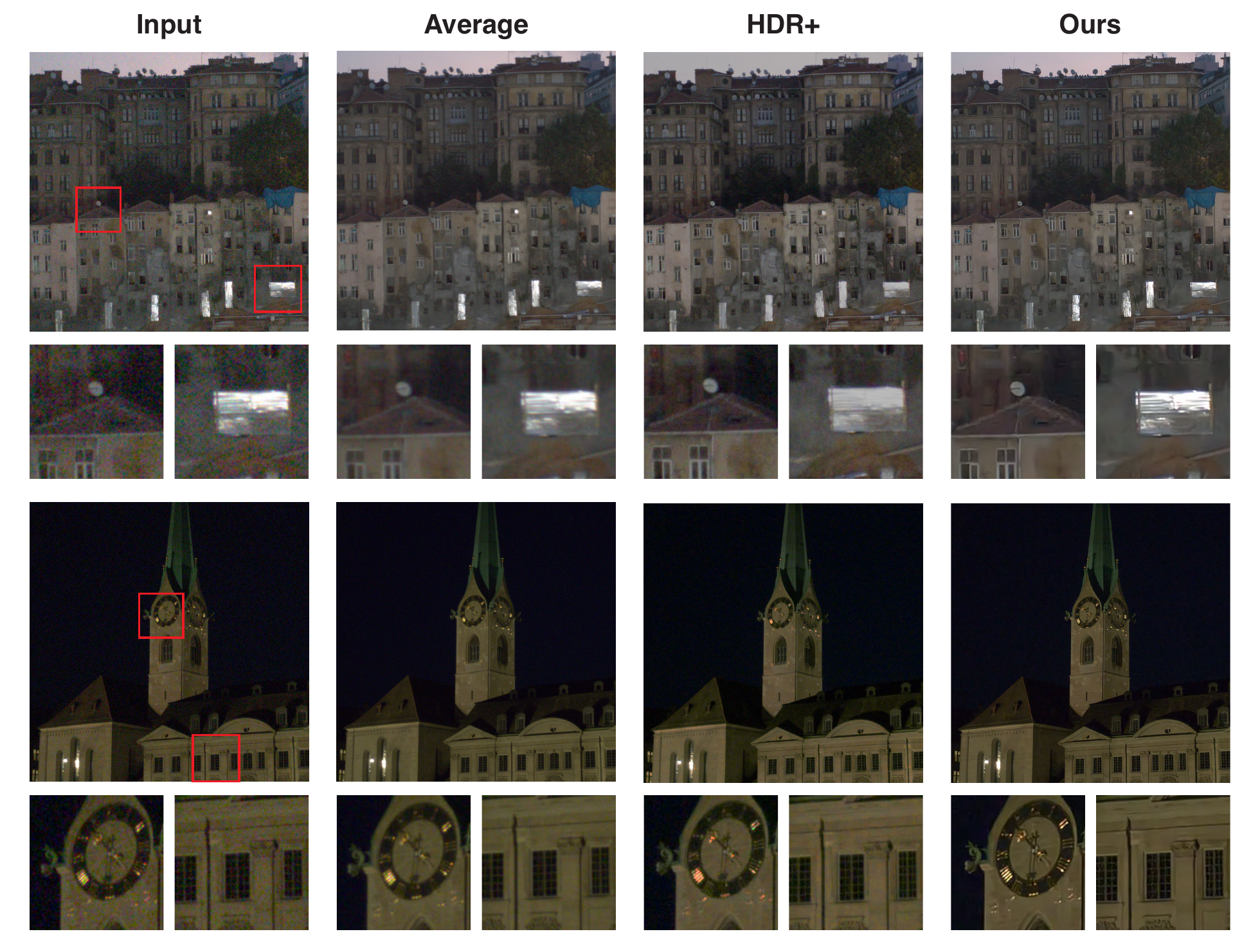}
\caption{\textbf{Denoising results on two real bursts on the HDR dataset}~\cite{hasinoff2016burst}. Our method produces a high level of denoising while keeping sharp details and maintains information in highlights.}\label{fig:hdrplus}
\end{figure*}

We evaluate our method on a held-out test set of Live Photos with synthetic additive white Gaussian noise added. In Table \ref{tab:nostab_comparison}, we compare our architecture with single frame models as well as the multi-frame method VBM4D~\cite{maggioni2011video, maggioni2012video}. We show qualitative results with $\sigma=50$ in Figure \ref{fig:sigma_50}.
In Figures \ref{fig:teaser} and \ref{fig:hdrplus} we demonstrate that our method is capable of denoising real sequences. This evaluation was performed on real noisy bursts from HDR+ \cite{hasinoff2016burst}. Please see our supplementary material for more results.

\begin{table*}[]
\centering
\scalebox{0.8}{
\begin{tabular}{|l|c|c|c||c|c|c|c|c|c|c|c||c|}
\hline
     & C2F   & C4F   &  C8F  & Ours 4L & \textbf{Ours 8L} & Ours 12L & Ours 16L & Ours 20L & Ours \textit{nostab} \\ \hline
PSNR & 30.89 & 31.83 & 32.15 & 33.01   & \textbf{33.62}   & 33.80    & 33.35    & 33.48    & 32.60       \\ \hline
\end{tabular}
}
\caption{\textbf{Ablation study on the Live Photos test sequences with additive white Gaussian Noise of $\sigma = 50$}. All models were trained on 8 frames long sequences. C2F, C4F and C8F represent \textbf{Concat} models which were trained on respectively 2, 4, and 8 concatenated frame as input. Ours \textit{nostab} was trained and tested on the unstabilized sequences.}
\label{tab:ablation}
\end{table*}

\subsubsection*{Ablation study}
We now evaluate our architecture choices, where we compare our full model, with 8 layers and trained on sequences of 8 frames with other variants.

\noindent{\bf Concat} We first compare our method with a naive multi-frame denoising approach, dubbed \textbf{Concat}, where the input consists of $n$ concatenated frames to a single pass denoiser.
We evaluated this architecture with $L = 20$ as well as $n = 2, 4$ and $8$. As we can see in Table~\ref{tab:ablation} this model performs significantly worse than our model.

\noindent{\bf Number of layers} We also evaluate the impact of the depth of the network by experimenting with $N = 4, 8, 12, 16$ and $20$. As can be seen in Figure~\ref{tab:ablation}, the 16 and 20 layers network fail to surpass both the 8 and 12 layers after 125 epochs of training, likely due to the increased depth and parameter count. While the 12 layers network shows a marginal 0.18dB increase over the 8 layer model, we decided to go with the latter as we did not think that the modest increase in PSNR was worth the $50\%$ increase in both memory and computation time.

\begin{figure}
  \centering
    \includegraphics[width=0.85\linewidth]{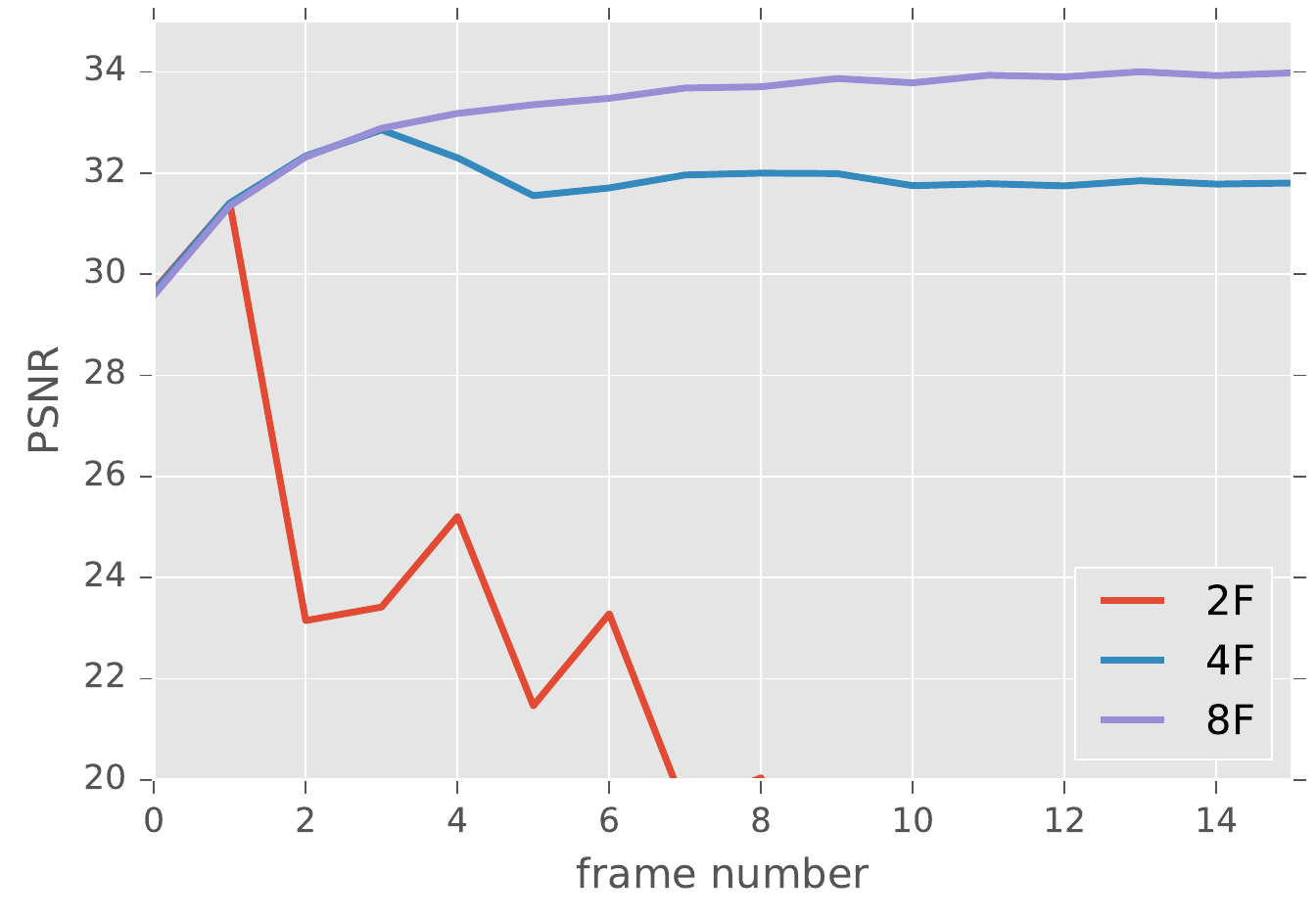}
    \caption{\textbf{Impact of the length $F$ of training sequences at test time.}
    We test 3 models which were trained with $F = 2, 4$ and $8$ on 16 frames-long test sequences.
    }
    \label{fig:frame_sweep_plot}
\end{figure}

\noindent{\bf Length of training sequences} Perhaps the most surprising result we encountered during training our recurrent model, was the importance of the number of frames in the training sequences. In Figure~\ref{fig:frame_sweep_plot}, we show that models trained on sequences of both 2 and 4 frames fail to generalize beyond their training length sequence. Only models trained with 8 frames were able to generalize to longer sequences at test time, and as we can see still denoise beyond 8 frames.

\noindent{\bf Pre-training} One of the main advantages of using a two-track network is that we can train the SFD track independently first. As mentioned just before, a sequence length of 8 is required to ensure generalization to longer sequences, which makes the training of the full model much slower than training the single-frame pass. As we show in Figure~\ref{fig:pretraining}, pre-training makes training the MFD significantly faster.

\iftrue
\begin{figure}
  \centering
    \includegraphics[width=0.85\linewidth]{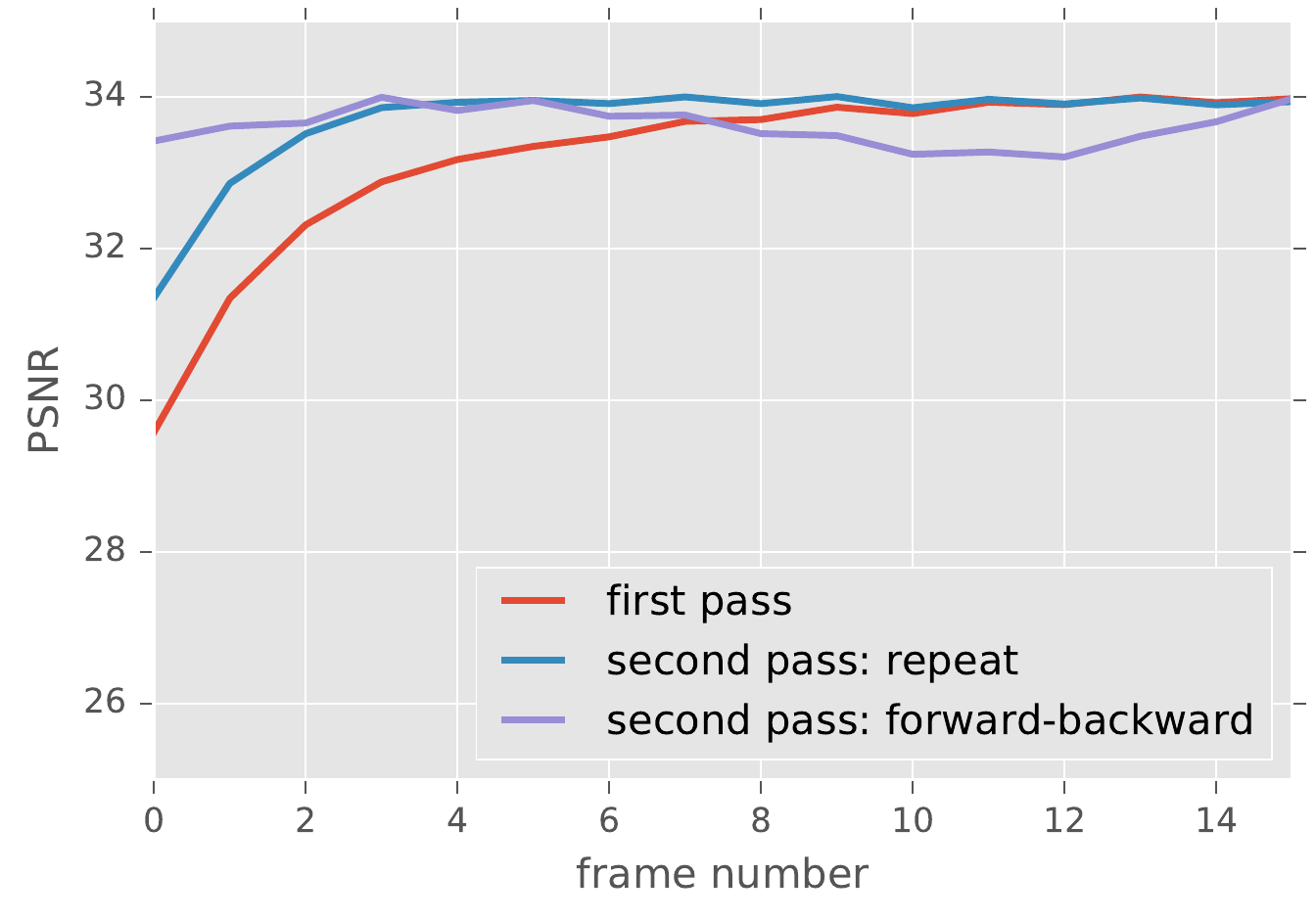}
    \caption{\textbf{Effect of frame ordering at test time.} We can see the burn-in period on the first pass (red) as well as on the repeat pass. Feeding the sequence forward, then backward, mostly alleviates this problem.}
    \label{fig:frame_ordering_plot}
\end{figure}
\fi

\noindent{\bf Frame ordering} Due to its recurrent nature, our network exhibits a period of burn-in, where the first frames are being denoised to a lesser extent than the later ones.
In order to denoise an entire sequence to a high quality level, we explored different options for frame ordering. As we show in Figure~\ref{fig:frame_ordering_plot}, by feeding the sequence twice to the network, we are able to obtain a higher average PSNR. We propose two variants, either \textbf{repeat} the sequence in the same order or reverse it the second time (named \textbf{forward-backward}). As we show in Figure~\ref{fig:frame_ordering_plot}, the forward-backward schedule does not suffer from burn-in nor flickering, thus meeting our 5\textsuperscript{th} goal. We thus use forward-backward for all our experiments.

\begin{figure*}
  \centering
    \newcommand{\turnheightnew}{0.24\textwidth}

\resizebox{0.60\textwidth}{!}{\begin{tabular}{@{\hskip 1mm}c@{\hskip 1mm}c@{\hskip 1mm}c@{\hskip 1mm}c@{\hskip 1mm}c@{\hskip 1mm}c@{}}

Bicubic & SFSR (Ours) & MFSR (Ours) & Ground truth \\

\includegraphics[width=\turnheightnew, trim=15cm 6cm 0cm 7cm, clip=true]{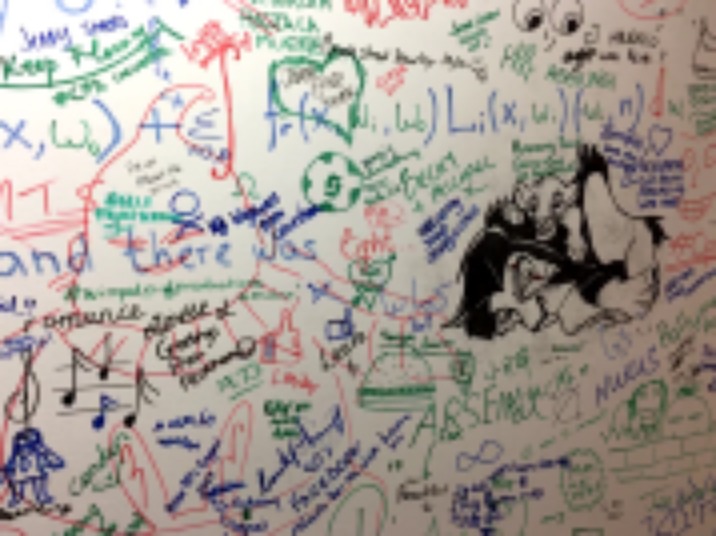} &
\includegraphics[width=\turnheightnew, trim=15cm 6cm 0cm 7cm, clip=true]{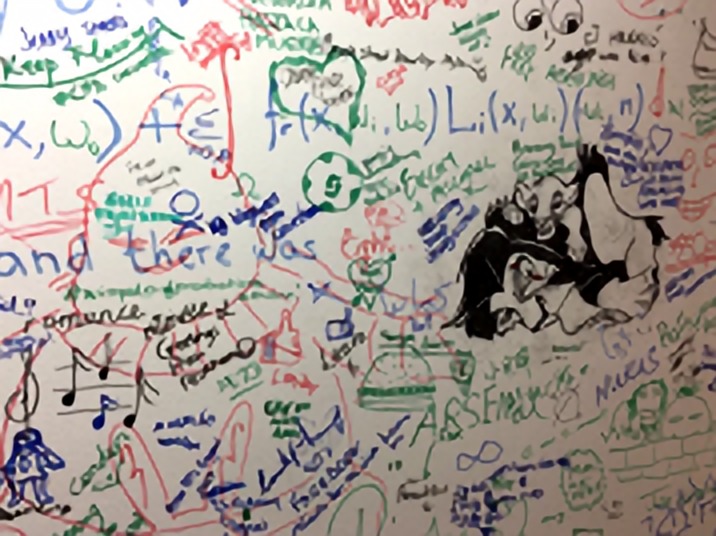} &
\includegraphics[width=\turnheightnew, trim=15cm 6cm 0cm 7cm, clip=true]{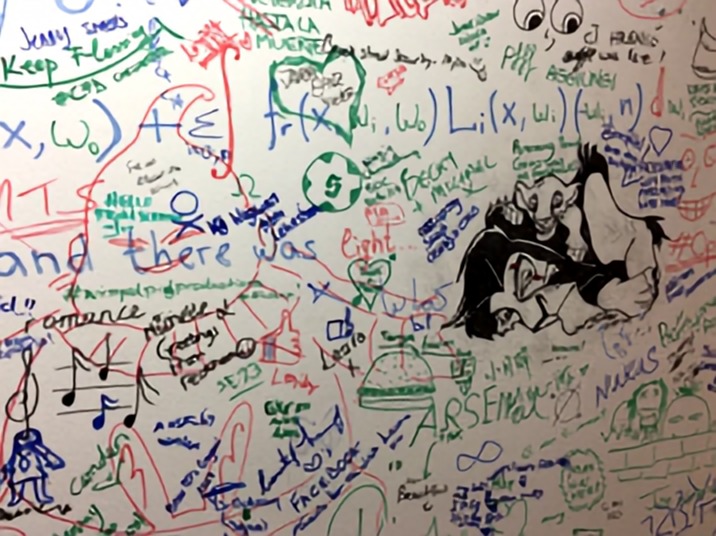} &
\includegraphics[width=\turnheightnew, trim=15cm 6cm 0cm 7cm, clip=true]{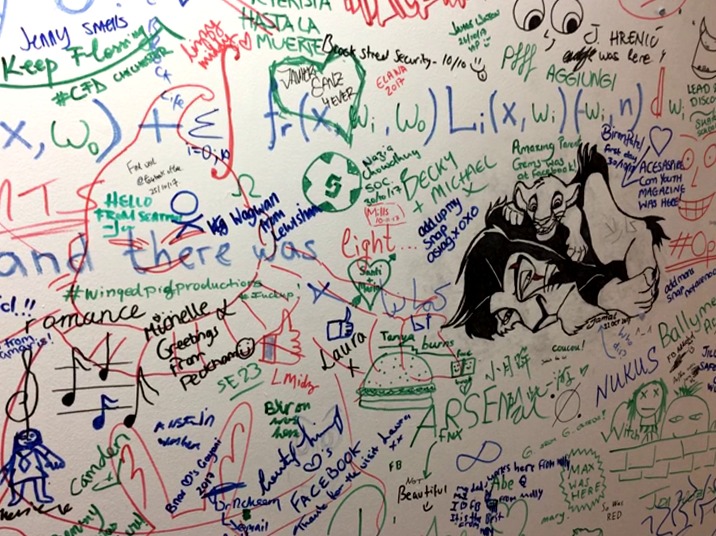}\\

\includegraphics[height=\turnheightnew, angle=90, trim=4cm 0cm 5cm 0cm, clip=true]{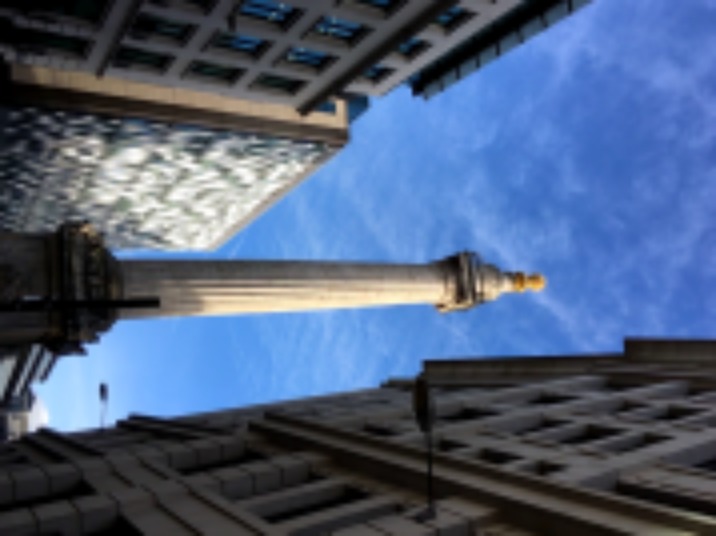} &
\includegraphics[height=\turnheightnew, angle=90, trim=4cm 0cm 5cm 0cm, clip=true]{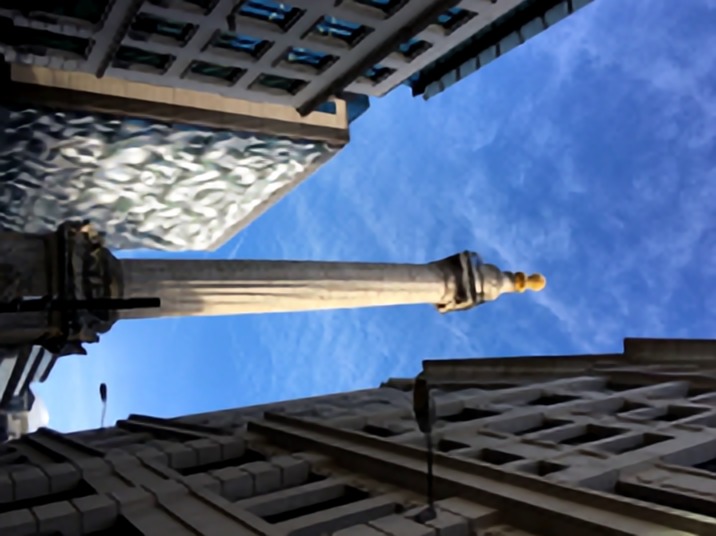} &
\includegraphics[height=\turnheightnew, angle=90, trim=4cm 0cm 5cm 0cm, clip=true]{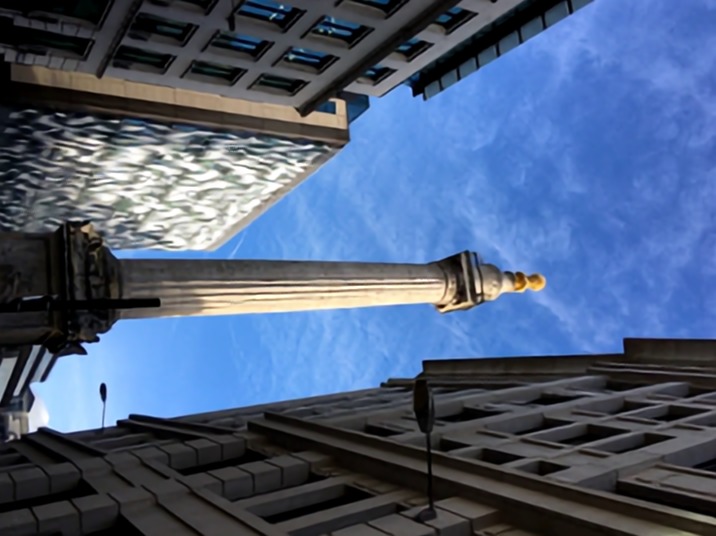} &
\includegraphics[height=\turnheightnew, angle=90, trim=4cm 0cm 5cm 0cm, clip=true]{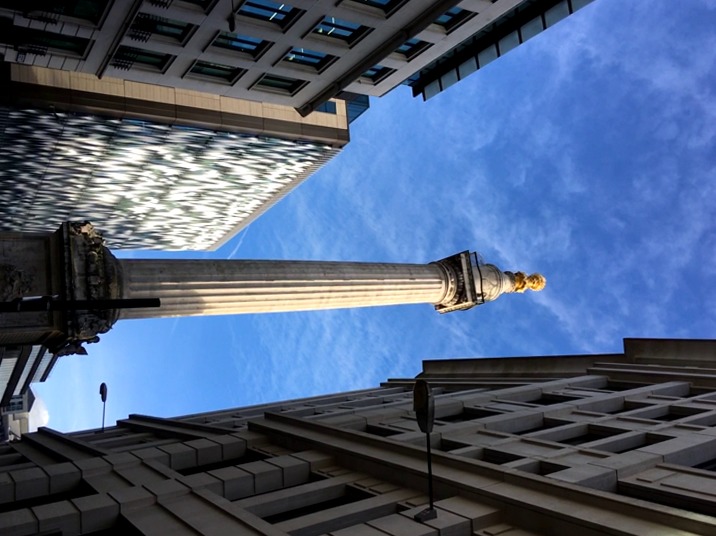}\\
\end{tabular}}
    \caption{\textbf{Multi-frame $4\times$ super-resolution on stabilized Live Photo test data.} While our single frame model achieves a good upsampling, the increase in sharpness from our multi-frame approach brings a significant quality improvement.}
    \label{fig:sres}
\end{figure*}

\begin{figure*}
\centering
\includegraphics[width=0.65\textwidth]{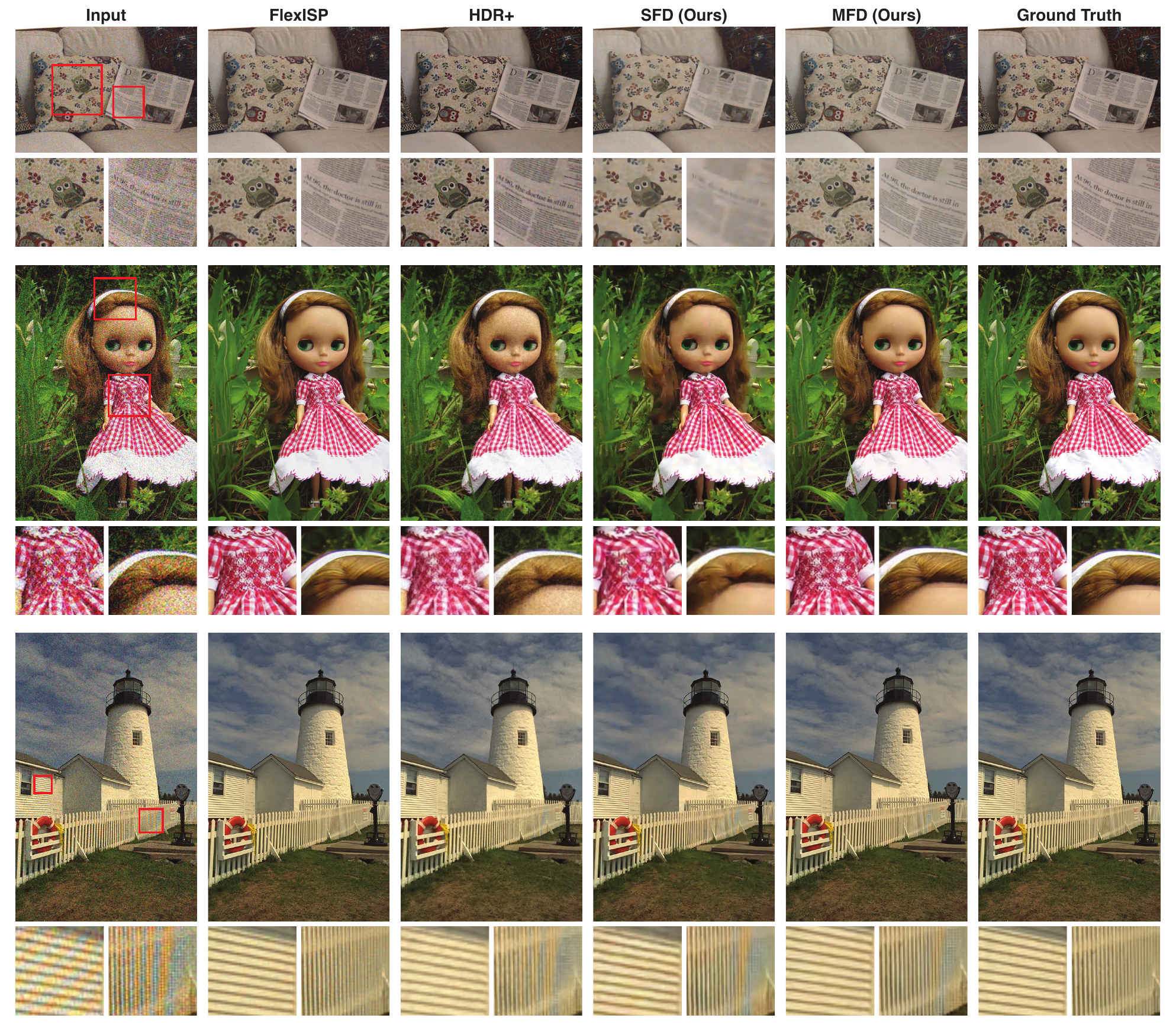}
\caption{\textbf{Denoising results on one real and two synthetic bursts on the FlexISP dataset}~\cite{heide2014flexisp}. From top to bottom: \textsc{livingroom}, \textsc{flickr doll} and \textsc{kodak fence}. Our recurrent model is able to match the quality of FlexISP on \textsc{flickr doll} and to beat it by 0.5dB on \textsc{kodak fence} despite showing demoisaicing artifacts.}\label{fig:flexisp}
\end{figure*}

\begin{table}[]
\centering
\scalebox{0.8}{
\begin{tabular}{|l|c|c|c|c|}
\hline
   & $\sigma=15$    &   $\sigma=25$ &    $\sigma=50$ &   $\sigma=75$  \\ \hline
BM3D       & 35.67 & 32.92 & 29.41 & 27.40 \\ \hline
VBM4D      & 36.42 & 33.41 & 29.14 & 26.60 \\ \hline
DnCNN      & 35.84 & 32.93 & 29.13 & 27.06 \\ \hline
DenoiseNet & 35.91 & 33.17 & 29.56 & 27.49 \\ \hline
\textbf{Ours} & \textbf{39.23} & \textbf{36.87} &\textbf{ 33.62} & \textbf{31.44} \\ \hline
\end{tabular}
}
\caption{\textbf{Multi-frame denoising comparison on Live Photo sequences}. Average PSNR for all frames on 1000 test 16-frames sequences with additive white Gaussian noise.}
\label{tab:nostab_comparison}
\vspace{-5mm}
\end{table}

\subsection{FlexISP}
We now compare our method with other denoising approaches on the FlexISP dataset and show our results in Figure~\ref{fig:flexisp}. Each sequence was denoised using the first 8 frames only. The synthetic sequences \textsc{flickr doll} and \textsc{kodak fence} were generated by randomly warping an input image and adding respectively additive and multiplicative white Gaussian noise of $\sigma = 25.5$, and additive with Gaussian noise of $\sigma = 12$ as well as simulating a Bayer filter.  We thus trained two models by replicating these conditions on our Live Photos dataset. On \textsc{flickr doll} our method achieves a PSNR of 29.39dB, matching FlexISP ($29.41$dB) but falling short of ProxImaL (30.23dB), not shown here. On \textsc{kodak fence} our recurrent model achieves a 0.5dB advantage over FlexISP (34.44dB) with a PSNR of 34.976dB.
Despite reaching a higher PSNR than FlexISP, our method does not mitigate the demosiacing artifacts on the fence, likely due to the absence of high frequency demosaicing artifacts in our training data.

\subsection{Super resolution}
To illustrate that our approach generalizes to tasks beyond denoising, and our 6\textsuperscript{th} goal, we trained our model to perform $4\times$ super-resolution, while keeping the rest of the training procedure identical to that of the denoising pipeline. Each input patch has been downsampled $4\times$, using pixel area resampling and then resized to their original size using bilinear sampling. Figure~\ref{fig:sres} shows a couple of our results. Please refer to the supplemental material for more results.

\section{Limitations}
Our single-frame architecture, based on \cite{remez2017deep, Zhang2017BeyondAG, Ledig_2017_CVPR}, makes use of stride-1 convolutions, enabling full-resolution processing across the entire network
They are however both memory and computationally expensive, and have a small receptive field for a given network depth. Using multiscale architectures, such as a U-Nets \cite{ronneberger2015u}, could help alleviate both issues, by reducing the computational and memory load, while increasing the receptive field. Finally while we trained our network on pre-stabilized sequences, we observed a significant drop in accuracy on unstabilized sequences, as can be seen in Table \ref{tab:ablation}, as well as unstability on longer sequences. It would be interesting to train the network to stabilize the sequence by warping inside the network such as in \cite{jaderberg2015spatial, Godard_2017_CVPR}.

\section{Conclusion}
We have presented a novel deep neural architecture to process burst of images.
We improve on a simple single frame architecture by making use of recurrent connections
and show that while single-frame models are reaching performance limits for denoising, our recurrent architecture vastly outperform such models for multi-frame data.
We carefully designed our method to align with the goals we stated in Section \ref{goals}. As a result, our approach achieves state-of-the-art performance in our Live Photos dataset, and matches existing multi-frame denoisers on challenging existing datasets with real camera noise.

{\small
\bibliographystyle{ieee}
\bibliography{biblio}
}

\end{document}